\documentclass{article}

\usepackage{arxiv}

\usepackage[utf8]{inputenc} 
\usepackage[T1]{fontenc}    
\usepackage{hyperref}       
\usepackage{url}            
\usepackage{booktabs}       
\usepackage{amsfonts}       
\usepackage{nicefrac}       
\usepackage{microtype}      
\usepackage{lipsum}
\usepackage{amsmath}
\newcommand\numberthis{\addtocounter{equation}{1}\tag{\theequation}}
\usepackage[export]{adjustbox}
\usepackage{fancyhdr}

\title{Causal Discovery using Compression-Complexity Measures}

\author{
Pranay SY, Nithin Nagaraj\\
Consciousness Studies Programme,\\ National Institute of Advanced Studies,\\ Indian Institute of Science Campus, Bengaluru, India. \\  \texttt{mail@pranaysy.com, nithin@nias.res.in  } \\
}

\begin{document}
\maketitle

\begin{abstract}
Causal inference is one of the most fundamental problems across all domains of science. We address the problem of inferring a causal direction from two observed discrete symbolic sequences $X$ and $Y$. We present a framework which relies on lossless compressors for inferring context-free grammars (CFGs) from sequence pairs and quantifies the extent to which the grammar inferred from one sequence compresses the other sequence. We infer $X$ causes $Y$ if the grammar inferred from $X$ better compresses $Y$ than in the other direction.
To put this notion to practice, we propose three models that use the Compression-Complexity Measures (CCMs) - Lempel-Ziv (LZ) complexity and Effort-To-Compress (ETC) to infer CFGs and discover causal directions without demanding temporal structures. We evaluate these models on synthetic and real-world benchmarks and empirically observe performances competitive with current state-of-the-art methods.
Lastly, we present two unique applications of the proposed models for causal inference directly from pairs of genome sequences belonging to the SARS-CoV-2 virus. Using a large number of sequences, we show that our models capture directed causal information exchange between sequence pairs, presenting novel opportunities for addressing key issues such as contact-tracing, motif discovery, evolution of virulence and pathogenicity in future applications.
\end{abstract}


\thispagestyle{fancy}
\section{Introduction}
The task of learning a causal model from observational data, or a combination of observational and interventional data, is commonly referred to as a causal discovery or causal structure learning \cite{weichwald_causality_2020}. Causal discovery from two variables based on observational data in the absence of time series or controlled interventions is a challenging problem and necessitates additional assumptions \cite{perner_semi-supervised_2018}. This is a ubiquitous problem in almost all domains of science, but particularly so in econometrics, meteorology, biology and medicine where interventional approaches are difficult or in several cases not feasible.

Model-free data-driven approaches for causal discovery have developed significantly over the past decade or so in an attempt to address the problem of causal discovery such as Granger Causality (GC) \cite{granger_investigating_1969}, Transfer Entropy (TE) \cite{schreiber_measuring_2000} and Compression-Complexity Causality (CCC) \cite{kathpalia_data-based_2019}. These methods have been used in various disciplines across neuroscience, climatology, econometrics, etc and rely on properties of time-series data. Such methods may have implicit assumptions that need to be met for satisfactory inference, and their robustness to different artefacts and nuisance variables typically depends on careful parameter calibration and selection for optimally accurate performance.

A class of model-free causal discovery methods do not assume a temporal structure in the data and are rooted in algorithmic information theory, chiefly based on the notion of Kolmogorov complexity. The Kolmogorov complexity of a finite binary string is the length of the shortest binary program that generates that string and reflects the computational resources needed to specify it. For two observed variables $X$ and $Y$, causal inference can be made by identifying the direction between $X$ and $Y$ where factorization of their joint distribution yields the lowest total Kolmogorov complexity \cite{budhathoki_origo_2018}. Since Kolmogorov complexity is not computable, it has typically been approximated using the Minimum Description Length (MDL) principle \cite{grunwald_minimum_2007}. MDL relies on lossless compressors to estimate the shortest program or description.

ORIGO is a method for causal inference based on the MDL principle which relies on tree-based compression \cite{budhathoki_origo_2018}. ORIGO infers that $X$ is a likely cause of $Y$ if better compression is achieved by first compressing $X$ and then compressing $Y$ given $X$, than in the other direction. The crux of this approach relies on the postulate that if $X$ causes $Y$, then describing $Y$ using $X$ will be easier or simpler than in the other direction \cite{pearl_2009}. In other words, if $X$ causes $Y$, $X$ will provide more information about $Y$ than vice versa. This approach is employed in ERGO, \cite{vreeken_causal_2015} which relies on estimates of complexity as proxies for Kolmogorov complexity instead of lossless compression. Both ORIGO and ERGO have been validated and benchmarked for univariate causal discovery.

Information theoretic quantities, such as Shannon entropy, suffer from several limitations for characterizing dynamical complexity of short and noisy time series data and better alternatives that capture complexity have been proposed and rigorously tested \cite{nagaraj_dynamical_2017}. These are called Compression-Complexity Measures (CCMs) and make use of lossless compression algorithms to characterize complexity, and has been validated using Lempel-Ziv (LZ) complexity \cite{lempel_complexity_1976} and Effort-to-Compress (ETC) algorithms \cite{nagaraj_new_2013}. Both LZ and ETC are grammar-based compression algorithms which infer a context-free grammar (CFG) based on an input sequence \cite{sayood_introduction_2017} and have found diverse applications spanning numerous domains. These CCMs have been demonstrated to be robust to noise, artefacts, missing data and further shown to capture complex behavior of dynamical and stochastic systems \cite{nagaraj_dynamical_2017}.

In the present paper, we propose a framework for causal discovery based on these CCMs that rely on inferred grammars to quantify information exchanged and shared between a pair of sequences. Based on this framework, we propose three models and evaluate their performance on synthetic as well as real-world benchmark data and compare their performance with that of ORIGO and ERGO.

Further, since our usage of CCMs within the proposed framework permits valid causal inference from sequences with different lengths, we propose and carry out unique applications of our models for assessing directional information exchange between genomic sequences of 16,619 SARS-CoV-2 virus isolates obtained from human samples. We believe this novel application has the potential to facilitate further investigations concerning key issues in bioinformatics, systems biology and epidemiology involving contact-tracing, epidemic monitoring, evolution and genetic interactions. 

\section{Materials and Methods}
We first propose our framework for univariate causal discovery from discrete symbolic sequences, which consist of ordered sets of elements or symbols, recorded with or without a concrete notion of time, for example: $\{2,3,1,2,3,2...\}$, $\{a,b,b,a,c,a,b,...\}$. We then present models based on this framework followed by empirical applications to synthetic and real-world data. We also present a novel experimental application of this framework for genome sequences.

\subsection{Framework}
Grammar-based compression schemes construct a context-free grammar (CFG) for a given string $x$ to be compressed by transforming $x$ into a CFG, $G$ \cite{sayood_introduction_2017}. The compressed representation of $x$ itself has been used for extracting information from and searching in sequences. We focus on $G$, which is a generative model of $x$ and contains its parsed structure based on rules relating to elements of $x$.

For two non-identical discrete symbolic sequences $x$ and $y$ with the same lengths and set of symbols, a lossless grammar-based compressor, $L$, will construct their CFGs, $G_x$ and $G_y$ respectively. $G_x$ represents an \textit{ideal} generative grammar for $x$, and $G_y$ for $y$. $L$ will output a compressed representation $L(x|G_x)$ or $L(x)$ of $x$ and analogously $L(y|G_y)$ or $L(y)$ of $y$. Since $x$ and $y$ are distinct sequences, $G_x$ and $G_y$ are also distinct. Using $G_x$ to parse or compress $y$, $L(y|G_x)$ leads to \textit{non-ideal} compression of $y$ compared to $L(y)$. Consequently, the compressed representation of $y$ can potentially be larger if compressed using $G_x$ instead of $G_y$. Such an approach in the context of the Minimum Description Length (MDL) principle \cite{grunwald_minimum_2007} has been employed for ORIGO \cite{budhathoki_origo_2018}.

The compressed representations of $x$ and $y$ can be described alternatively in terms of Compression-Complexity Measures (CCMs) \cite{nagaraj_dynamical_2017} besides description lengths. CCMs are measures of complexity derived from lossless data compression algorithms and have been shown to be robust estimators of the complexity of discrete as well as continuous dynamical ans stochastic systems. We use CCMs to measure the complexity of information exchanged between pairs of sequences based on their estimated  generative grammars.

In the present context, we argue that the CCM of $L(y|G_x)$ is different from the CCM of $L(y)$. $G_x$ may compress $y$ better or poorer than $G_y$, resulting in CCM of $L(y|G_x)$ lesser or greater than that of $L(y)$ respectively. Similarly estimating CCMs of $L(x|G_y)$ and $L(x)$ and comparing both directions permits an examination of the influence of $G_x$ and $G_y$ on the sequences $x$ and $y$. Learning $G_x$ for compressing $y$ first will be better than learning $G_y$ for compressing $x$, if $x$ causes $y$. Under this assumption of `cause subsumes effect' in terms of grammar complexity, and the assumption that $x$ and $y$ don't share a common cause, we extend this notion further to quantify learning in terms of complexity using CCMs. We present two distinct formulations for causal discovery, one based on \textit{penalty} and another based on \textit{efficacy} of compression using \textit{non-ideal} grammars.

\subsubsection{Penalty}
The difference CCM$_L(y|G_x) - $CCM$_L(y)$ represents the cost in terms of compression-complexity incurred by compressing $y$ using a \textit{non-ideal} grammar, in this case $G_x$. The better $G_x$ can compress $y$, the smaller the CCM of $L(y|G_x)$ and thus, smaller the cost. This cost or penalty, $P$, is directional and consists of the penalty of compressing $y$ using $G_x$, $P_{x\rightarrow y}$ as well as the penalty of compressing $x$ using $G_y$, $P_{y\rightarrow x}$.
\begin{align*}
P_{x\rightarrow y} &= \text{CCM}_L(y|G_x) - \text{CCM}_L(y), \\
P_{y\rightarrow x} &= \text{CCM}_L(x|G_y) - \text{CCM}_L(x). \numberthis \label{eqn1}
\end{align*}
If penalty in one direction is lesser than that in the other direction, then one sequence's grammar can better compress the other sequence. This implies that the if $P_{x\rightarrow y} < P_{y\rightarrow x}$ then the inferred generative model $G_x$ can account for $y$ better than $G_y$ can account for $x$. We extend this to arrive at the following rules for inferring a causal direction:
\begin{align*}
\text{If } P_{x\rightarrow y} &< P_{y\rightarrow x} \text{, we infer } x\rightarrow y, \\
\text{If } P_{x\rightarrow y} &> P_{y\rightarrow x} \text{, we infer } y\rightarrow x, \\
\text{If } P_{x\rightarrow y} &= P_{y\rightarrow x} \text{, we are undecided. }
\end{align*}
A threshold can be introduced for the differences for inferring direction in practice. The difference $|P_{y\rightarrow x} - P_{x\rightarrow y}|$ is an indication of the causal strength or strength of causal evidence in favor of the inferred direction, the larger the difference, the stronger the causal evidence in the inferred direction.

\subsubsection{Efficacy}
The difference CCM$_L(y|G_x) - $CCM$_L(x)$ represents the efficacy of $G_x$ towards complete lossless compression of $y$. If $G_x$ compresses $y$ effectively, then \textit{residual y} will have lower CCM and this difference will be small. The difference can also be viewed as the additional compression-complexity introduced by using $G_x$ to compress $y$ instead of using $G_x$ to compress $x$. This efficacy, $E$ is directional and consists of the efficacy of compressing $y$ using $G_x$, $E_{x\rightarrow y}$ as well as the efficacy of compressing $x$ using $G_y$, $E_{y\rightarrow x}$.
\begin{align*}
E_{x\rightarrow y} &= \text{CCM}_L(y|G_x) - \text{CCM}_L(x), \\
E_{y\rightarrow x} &= \text{CCM}_L(x|G_y) - \text{CCM}_L(y). \numberthis \label{eqn2}
\end{align*}
If efficacy in one direction is greater than that in the other direction, then one sequence's grammar can more effectively compress the other sequence. This implies that the if $E_{x\rightarrow y} > E_{y\rightarrow x}$ then the inferred generative model $G_x$ can account for $y$ better than $G_y$ can account for $x$. We extend this to arrive at the following rules for inferring a causal direction:
\begin{align*}
\text{If } E_{x\rightarrow y} &> E_{y\rightarrow x} \text{, we infer } x\rightarrow y, \\
\text{If } E_{x\rightarrow y} &< E_{y\rightarrow x} \text{, we infer } y\rightarrow x, \\
\text{If } E_{x\rightarrow y} &= E_{y\rightarrow x} \text{, we are undecided. }
\end{align*}
As with the penalty approach, a threshold can be introduced for the differences for inferring direction here as well. The difference $|E_{y\rightarrow x} -E_{x\rightarrow y}|$ indicates causal strength or strength of causal evidence in favor of the inferred direction, the larger the difference, the stronger the causal evidence in the inferred direction.

\subsection{Models}
The framework presented involves describing a sequence and its compressed representation using CCMs. While various lossless grammar-based compression schemes exist, CCMs have been studied and validated using Lempel-Ziv (LZ) complexity and Effort-To-Compress (ETC) measures \cite{nagaraj_dynamical_2017}. We present implementations of our framework using these two CCMs for both the \textit{penalty} and \textit{efficacy} approaches. We also use and describe the ORIGO and ERGO models from two similar frameworks rooted in algorithmic information theory, for comparative assessment of performance.

\subsubsection{ETC-based}
ETC is defined as the effort to compress \cite{nagaraj_new_2013} an input symbolic sequence using the lossless compression algorithm known as Non-Sequential Recursive Pair Substitution (NSRPS) \cite{ebeling_grammars_1980}. Numerically ETC is an estimate of the number of steps required by NSRPS to compress an input sequence to a constant sequence or a sequence with zero entropy. ETC has been demonstrated to reliably capture complexity of short and noisy time series and is robust to missing data \cite{nagaraj_lossless_2011, nagaraj_dynamical_2017}. The algorithm is functionally similar to the Re-Pair lossless compressor \cite{larsson_off-line_2000, calcagnile_non-sequential_2010} although ETC has found broad applications in diverse domains including signal processing, de-noising, cognitive studies, heart-rate variability, Schizophrenia research, etc \cite{balasubramanian_chaos_2020, balasubramanian_vagus_2017, funkhouser_role_2020, li_noise_2018, kiefer_shaping_2020, virmani_novel_2019, thanaj_analysis_2018}. ETC has also been employed as the compression-complexity measure underlying the Compression-Complexity Causality (CCC) framework for robust causal inference through data-based intervention \cite{kathpalia_data-based_2019}.
  
\textit{Penalty approach}  

Since ETC is an off-line compressor, $G_x$ as well as $G_y$ can be stored separately and \textit{residual y} and \textit{residual x} can be observed individually, simplifying the formulation to:
\begin{align*}
\text{ETC-}P_{x\rightarrow y} &= \text{ETC}(y|G_x) + \text{ETC}(y_{\text{residual}}) - \text{ETC}(y), \\
\text{ETC-}P_{y\rightarrow x} &= \text{ETC}(x|G_y) + \text{ETC}(x_{\text{residual}}) - \text{ETC}(x). \numberthis \label{eqn3}
\end{align*}

\textit{Efficacy approach}  

The efficacy of $G_x$ for compressing $y$ and vice-versa can be assessed by directly estimating CCMs of the residual sequences, leading to the formulation:
\begin{align*}
\text{ETC-}E_{x\rightarrow y} &= \lambda \text{ETC}(y_{\text{residual}}), \\
\text{ETC-}E_{y\rightarrow x} &= \lambda \text{ETC}(x_{\text{residual}}). \numberthis \label{eqn4}
\end{align*}
where $\lambda$ is a normalization factor to allow for better comparability of ETC estimates of the \textit{residuals}.

We use the standard normalization factor for ETC where $\lambda=(\text{length}(y_{\text{residual}})-1)^{-1}$, referred to as the ETC-E formulation.

\subsubsection{LZ-based}
Lempel-Ziv complexity is defined as the number of different sub-strings encountered as the binary sequence is viewed as a stream (from left to right) \cite{lempel_complexity_1976}. LZ complexity has been used extensively in a wide spectrum of domains from linguistics, phylogenetics and neural-spike train analysis to mechanical fault identification and siezure detection \cite{li_intelligent_2020, yakovleva_eeg_2020,szczepanski_characterizing_2004,yu_viral_2014,pregowska_using_2019}. We demonstrate the usage of LZ complexity in the penalty model, where $\text{CCM}_L(y|G_x)$ is estimated as the \textit{`joint'} LZ complexity of $x$ and $y$, or $\text{LZ}(x,y)$.

As a CCM, LZ complexity is an estimator of the entropy rate of sequences generated from stochastic dynamical systems \cite{nagaraj_dynamical_2017}. We use this relation between LZ complexity and entropy to formulate a measure of penalty based on conditional entropy, defined as $H(y|x)=H(x,y)-H(x)$, where $H(x,y)$ is the joint entropy of $x$ and $y$. The analogous expression $\text{LZ}(y|x) = \text{LZ}(x,y) - \text{LZ}(x)$ represents the excess complexity inherent in the \textit{`joint'} compression-complexity of $x$ and $y$ taken together compared to that of $x$ alone. This excess complexity corresponds to the \textit{penalty} incurred when compressing $y$ using $x$. If $x$ causes $y$ then the penalty in the direction $x\rightarrow y$ will be lesser than that in the direction $y\rightarrow x$, leading to the following formulation:
\begin{align*}
\text{LZ-}P_{x\rightarrow y} &= \text{LZ}(x,y) - \text{LZ}(x), \\
\text{LZ-}P_{y\rightarrow x} &= \text{LZ}(x,y) - \text{LZ}(y). \numberthis \label{eqn5}
\end{align*}
Different approaches for estimating $\text{LZ}(x,y)$ exist, such as the vectorial form of joint LZ complexity \cite{ZOZOR2005285} which evaluates index-matched pairs of symbols from both $x$ and $y$ synchronously, as well as the concatenated forms $\text{LZ}(xy)$ and $\text{LZ}(yx)$ used for estimating compression distances in bioinformatics \cite{makinen_11_2015}. We use the concatenation approach for estimating ${\text{LZ}}(x,y)$ with the caveat that for long sequences $\text{LZ}(xy)$ and $\text{LZ}(yx)$ result in identical estimates, however for short sequences this symmetry may be violated.

\subsubsection{ORIGO}
ORIGO is an efficient method for causal inference \cite{budhathoki_origo_2018} from binary data based on the minimum description length (MDL) principle \cite{grunwald_minimum_2007}. It relies on the tree-based lossless compressor, PACK \cite{tatti_finding_2008}, for searching a set of decision trees that encode the data most succinctly. ORIGO works with both univariate and multivariate data with an equal number of observations and has been extensively evaluated previously. We chose ORIGO for comparison since it works directly on the data and does not require assumptions about distributions, similar to our proposed framework.

\subsubsection{ERGO}
ERGO is a robust method for causal inference based on direction of information and uses cumulative and Shannon entropy \cite{vreeken_causal_2015}. It determines the amount of information one set of data provides about another and vice versa, and infers causal direction based on the strongest direction of information. ERGO also works with both univariate and multivariate data and has been shown to be robust to noise. We chose ERGO for comparison due to the principles of causal inference it shares with our proposed framework based on direction of information. While we employ robust CCMs to estimate strengths under the penalty and efficacy models, ERGO uses conditional cumulative entropy to estimate relative amount of information shared based on Kolmogorov Complexity.

\subsection{Implementation}
An open-source implementation of ETC in Python (version 3.8) was used, with some steps of NSRPS implemented in Cython \cite{behnel_cython_2011} and NumPy \cite{harris_array_2020}, available on GitHub: \\https://github.com/pranaysy/ETCPy/\\ Simulations and processing of data used in the paper were carried out using Python scripts. LZ complexity was computed using the implementation provided in the open-source EntroPy package for Python:\\https://github.com/raphaelvallat/entropy\\Estimates for ORIGO and ERGO were computed using the reference open-source implementation in Python and Java respectively, available publicly:\\http://eda.mmci.uni-saarland.de/prj/origo/

\subsection{Synthetic unidirectional coupling}
The autoregressive (AR) model is widely used in statistics, econometrics as well as signal processing ans is used to describe time-varying processes that linearly depend on their own past as well as the past of other processes \cite{shumway_time_2017}. The AR model is commonly used for tests of causal inference and is foundational to the Granger causal framework \cite{granger_investigating_1969}. We used the AR model for unidirectional causal inference and simulated autoregressive processes of order one (AR(1)) as follows with $X$ and $Y$ as the dependent and independent processes respectively. 
\begin{align*}
X(t) &= aX(t-1) + \phi Y(t-1) + \epsilon_{X,t}, \\
Y(t) &= bY(t-1) + \epsilon_{Y,t}. \numberthis \label{eqn6}
\end{align*}
where $a=b=0.8$, $t=1$ to 1,000s, sampling period $=1$s. We varied the coupling parameter, $\phi$, from 0 to 0.95 in steps of 0.05 for a total of 20 values. Noise terms $\epsilon_X, \epsilon_Y=vN$, where noise intensity, $v=0.01$ and follows the standard normal distribution. For each value of $\phi$, we randomly sampled 1,000 trials of $X$-$Y$ pairs, and discretized each sampled $X$-$Y$ pair to binary sequences using an equi-width binning strategy. We then estimated causal direction and scores using all five models for each trial, for a total of 20,000 trials.

We evaluated performance as accuracy against decision rates across varying coupling strengths for each model. We calculated the absolute differences in causal influences in both directions, $x\rightarrow y$ and $y\rightarrow x$, and sorted them in descending order. We then calculated the accuracy of causal discovery over the top $k$\% differences, where $k$ is the decision rate, to assess confidence of the models in the inferred direction. For pairs where a direction could not be inferred, we flip a fair coin, similar to the methodology previously used for ORIGO \cite{budhathoki_origo_2018}. We also present overall accuracy based on 1,000 trials for each coupling strength for all five models along with estimates of area under the Receiver Operating Characteristic curve (AUROC) and area under the Precision-Recall curve (AUPRC).

\subsection{Real-world benchmarks}
We evaluated the performance of all five models on benchmark cause-effect pairs with known ground truth, which consists of real-world causal variables \cite{mooij_distinguishing_2015}. The dataset consists of both multivariate and univariate cause-effect pairs, and here we consider only the 90 univariate ones. In order to facilitate comparability, we used the interaction-preserving discretization (IPD) strategy \cite{nguyen_unsupervised_2014} for all the pairs as used previously \cite{budhathoki_origo_2018}. Similar to the evaluation of performance for synthetic data, we investigated accuracy against decision rates as well as AUROC and AUPRC for all five models.

\subsection{Causal discovery in genome sequences}
The highly contagious pathogen called Severe Acute Respiratory syndrome coronavirus 2 (SARS-CoV-2) was first identified in December 2019 \cite{coronaviridae_study_group_of_the_international_committee_on_taxonomy_of_viruses_species_2020} as the cause of a respiratory illness designated coronavirus disease 2019, or Covid-19, classified as a global pandemic by the World Health Organization (WHO) on 11th March 2020. Coronaviruses are generally well known to evolve environmentally and have high virulence conferring effective transmission and immune evasion strategies \cite{chen_pathogenicity_2020}. At the time of writing (October 15th, 2020) over 38 million cases and 1.09 million fatalities have been reported globally \cite{covid_global_nodate}, with neither a cure for Covid-19 nor an effective vaccine within sight \cite{beigel_remdesivir_2020}.

Effective countermeasures against the virus require the development of data and tools to understand and monitor its spread and immune responses to it \cite{salvamani_understanding_2020}. Methods grounded in information theory and data compression have found numerous applications in genomics - global sequence analysis, phylogenetics, evolutionary modeling, sequence complexity, motif discovery and classification, analysis of secondary structures, etc \cite{vinga_information_2014, nalbantoglu_data_2009}. Recently, approaches for causal inference that rely on graphical models \cite{glymour_review_2019} as well as additive noise models (ANMs) \cite{hoyer2009nonlinear} have received attention for causal genetic analysis and bivariate causal discovery from gene expression data \cite{jiao_bivariate_2018}.

Here, we present two unique applications of our proposed grammar-based framework for inferring directions and strengths of causal information exchange directly from genome sequences obtained from viral isolates of SARS-CoV-2. In experiment 1 we assess directions of causal interaction between genome sequences using the three models presented. In experiment 2 we investigate the use of causal strengths for classification of causal sequences from known candidates.

\subsubsection{Experiment 1}
We hypothesized that the SARS-CoV-2 consensus sequence \cite{wang_establishment_2020} would capture information representative of the initial state of the viral genome prior to the global outbreak. Subsequent transmission of the virus would be accompanied with genome evolution over time, altering the nucleotide sequence's information content. Since this evolved sequence contains changes accumulated over the consensus sequence, the evolved sequence can be viewed as a derivative of the consensus sequence, admitting a direction. Alternatively then, the consensus sequence potentially `causes' the evolved sequence, permitting us to examine this within the causal discovery framework and test if our models indeed identify this hypothesized direction. Similarly, we also examined if the first sequence reported in each country `caused' subsequent sequences in that country. Since individual nucleotide sequences do not contain temporal structures to suggest an ordering in time, these hypotheses of causal information exchange rely on an external temporal sequence of events suggested by disease spread and genome evolution.

To evaluate this hypothesis, we computed estimates from the three proposed models of our framework - ETC-E, ETC-P and LZ-P. Since pairs of sequences are never evaluated simultaneously at a symbol-level under these three models, novel applications to genomic data are possible where sequences with different lengths are the norm. Pairwise estimates were computed for sequences from each country such that one sequence of a pair was first fixed to the consensus sequence, and later fixed to the first sequence for that country, resulting in estimates of direction and strength. Based on these estimates, we evaluated the hypothesis involving causal directions using the proportion of sequences for each country for which causal discovery was made in the expected direction for both the consensus sequence as well as the first sequence per country.

These experiments involved estimation of causal directions for pairs of genome sequences with variable lengths. A total of 16,619 high quality complete nucleotide sequences obtained from human hosts from 19 countries were obtained - 12,556 from the GISAID initiative's EpiCoV database \cite{elbe_data_2017} and the remaining from GenBank \cite{10.1093/nar/gks1195}. A complete table of the sequences used, with accession identifiers, is provided in supplement S1. Each sequence was encoded numerically with the following mapping: A=1, C=2, G=3, T=4. Sequences with ambiguous nucleotides were not subject to analysis. 

\subsubsection{Experiment 2}
We focused on the problem of identifying and classifying causal origins of observed genomic sequences based on known candidate sequences. We consider genome sequences from SARS-CoV-2 isolates and test the hypothesis that the strength of causal influence on these sequences from the SARS-CoV-2 consensus sequence would be greater than the SARS-CoV-1 consensus sequence. Since SARS-CoV-1 and SARS-CoV-2 are closely related \cite{coronaviridae_study_group_of_the_international_committee_on_taxonomy_of_viruses_species_2020}, through this experiment we attempted to assess the sensitivity of causal strengths estimated from the three proposed models towards distinguishing causal influences from the two strains.

Six countries were randomly chosen from the dataset used in experiment 1 and from each country, subsets of genome sequences were randomly chosen. The number of sequences sampled per country were: China - 96, France - 24, India - 44, Netherlands - 98, Russia - 92 and United States - 84. A list of identifiers and accession IDs for these sequences is provided in the supplement S2. Causal strengths were estimated from each of the three models by fixing one sequence to the SARS-CoV-1 (GenBank Accession ID: NC\_004718.3) or SARS-CoV-2 (GenBank Accession ID: NC\_045512.2) consensus sequence and the other to one of the sequences from the database. The direction of causal influence inferred by the causal models was ignored as this direction was fixed \textit{a priori} based on the two candidate sequences.

Statistical analysis was carried out in the R statistical computing environment \cite{rcore} using functions from Rand R. Wilcox's robust statistics library \cite{wilcox_introduction_2016}. We tested for differences in the strength of causal influence from SARS-CoV-2 and  SARS-CoV-1 consensus sequences using resampling-based robust statistics. The bootstrap-t method was used to compare 20\% trimmed means of causal strengths from SARS-CoV-1 and SARS-CoV-2 using the function \textit{yuenbt}. Symmetric 95\% confidence intervals were computed based on 5,000 bootstrap iterates. Each test between the two strains per country and model was treated independently and correction for multiple comparisons was not carried out. All computed statistics are provided in supplement S3.

\begin{figure*}
	\centering
		\includegraphics[scale=0.95, center]{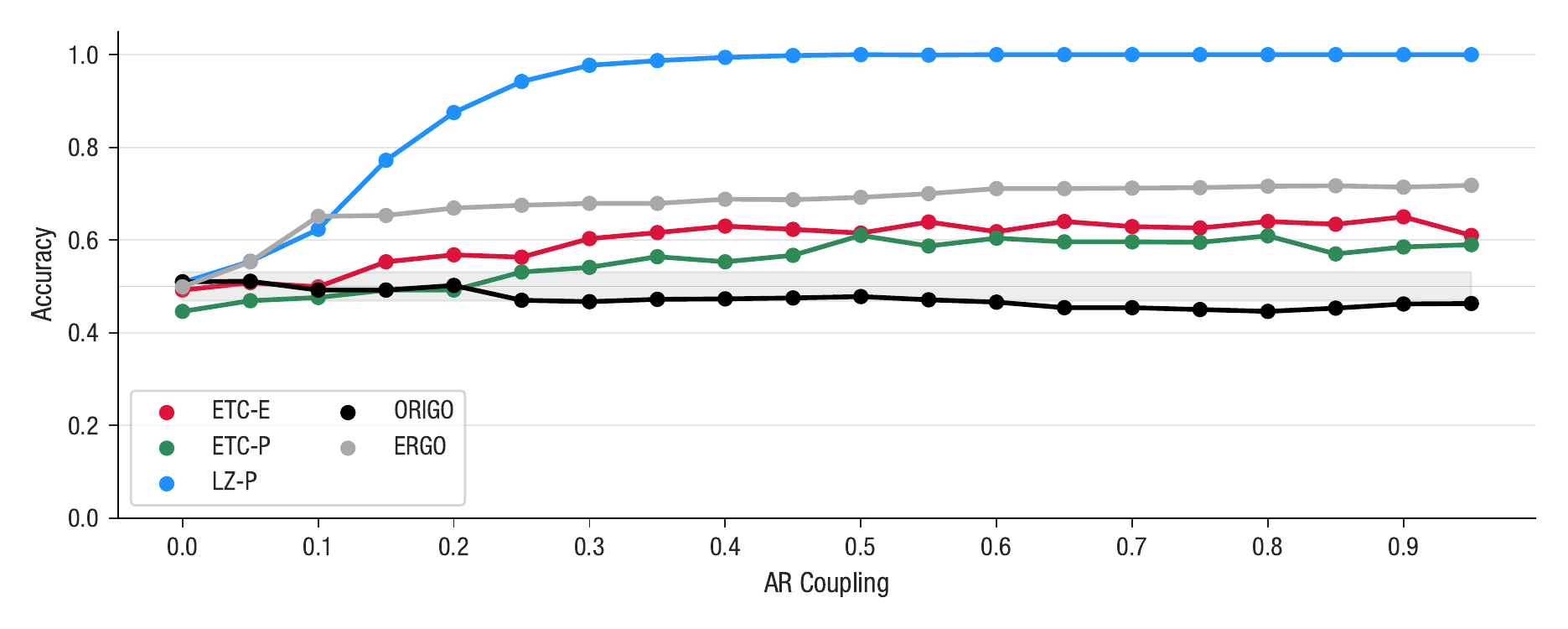}
	\caption{Overall accuracy of the five models across strengths of AR coupling.}
	\label{FIG:1}
\end{figure*}

\begin{figure}
	\centering
		\includegraphics[scale=0.95, center]{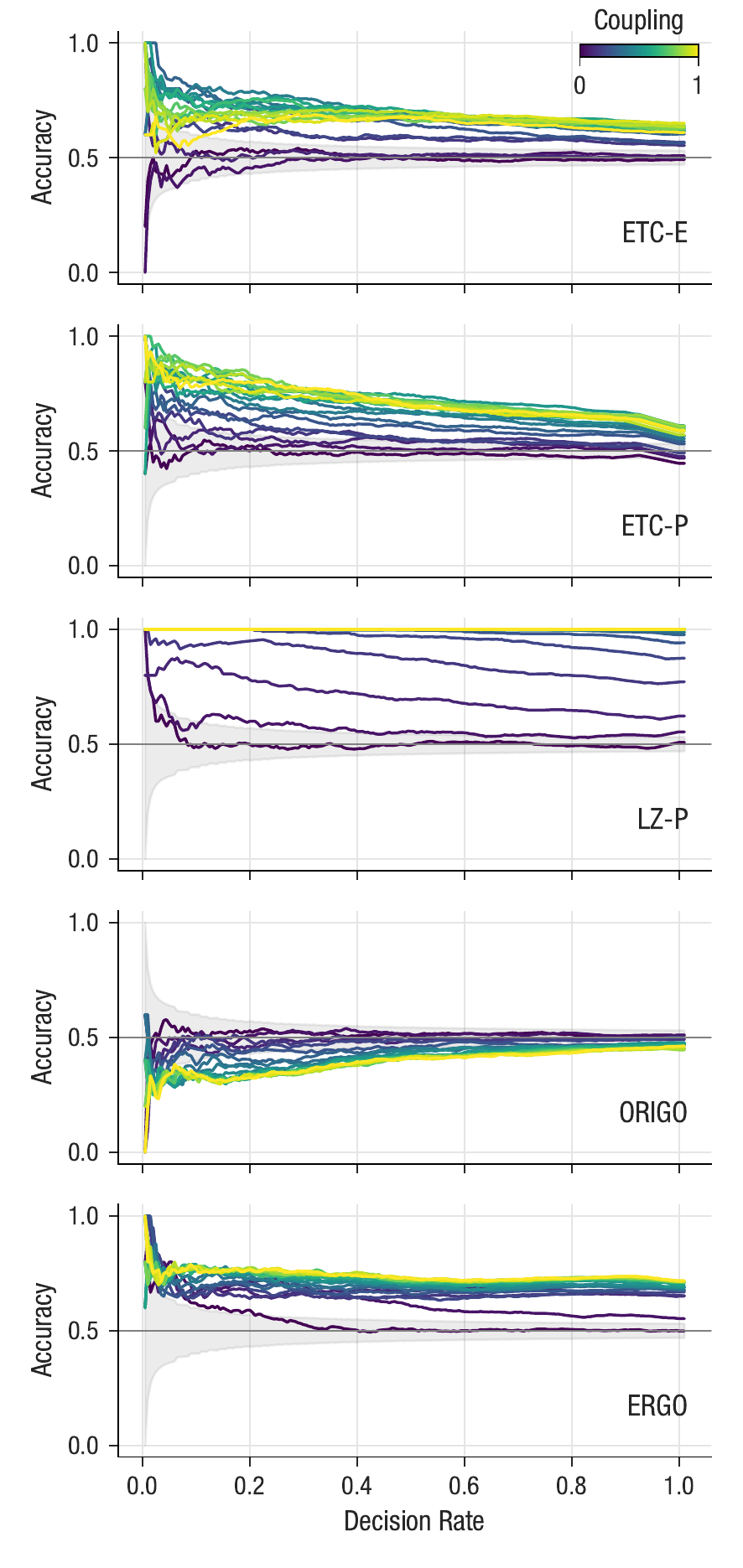}
	\caption{Accuracy of the five models across decision rates for varying strengths of AR coupling. Gray band indicates 95\% CI of the binomial distribution.}
	\label{FIG:2}
\end{figure}

\begin{figure}
	\centering
		\includegraphics[scale=0.95, center]{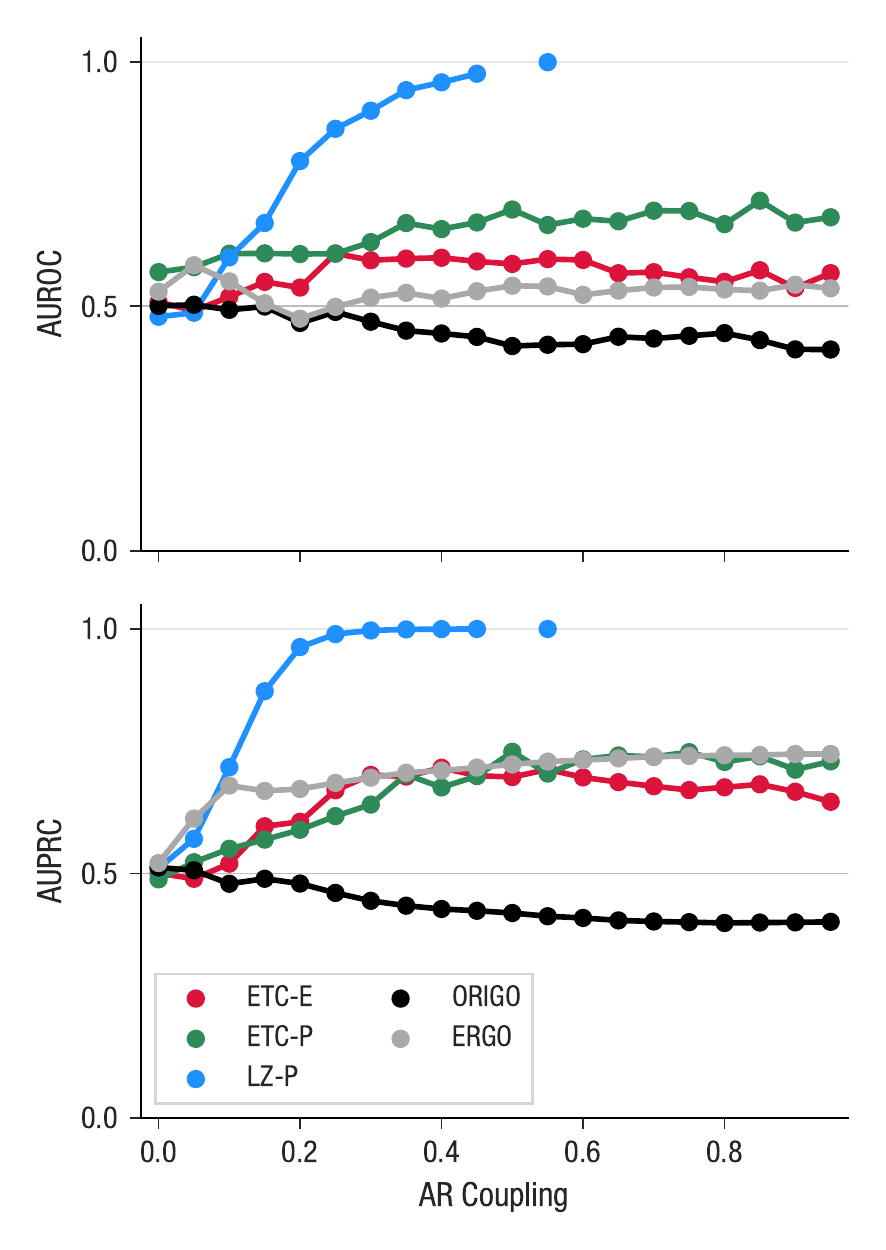}
	\caption{Area under Receiver Operating Characteristic (AUROC) and Area under Precision-Recall Curve (AUPRC) for the five models across strengths of AR coupling. In the case of LZ-P, the AUROC and AUPRC are undefined for 100\% accuracy.}
	\label{FIG:3}
\end{figure}

\begin{figure*}
	\centering
		\includegraphics[scale=0.95, center]{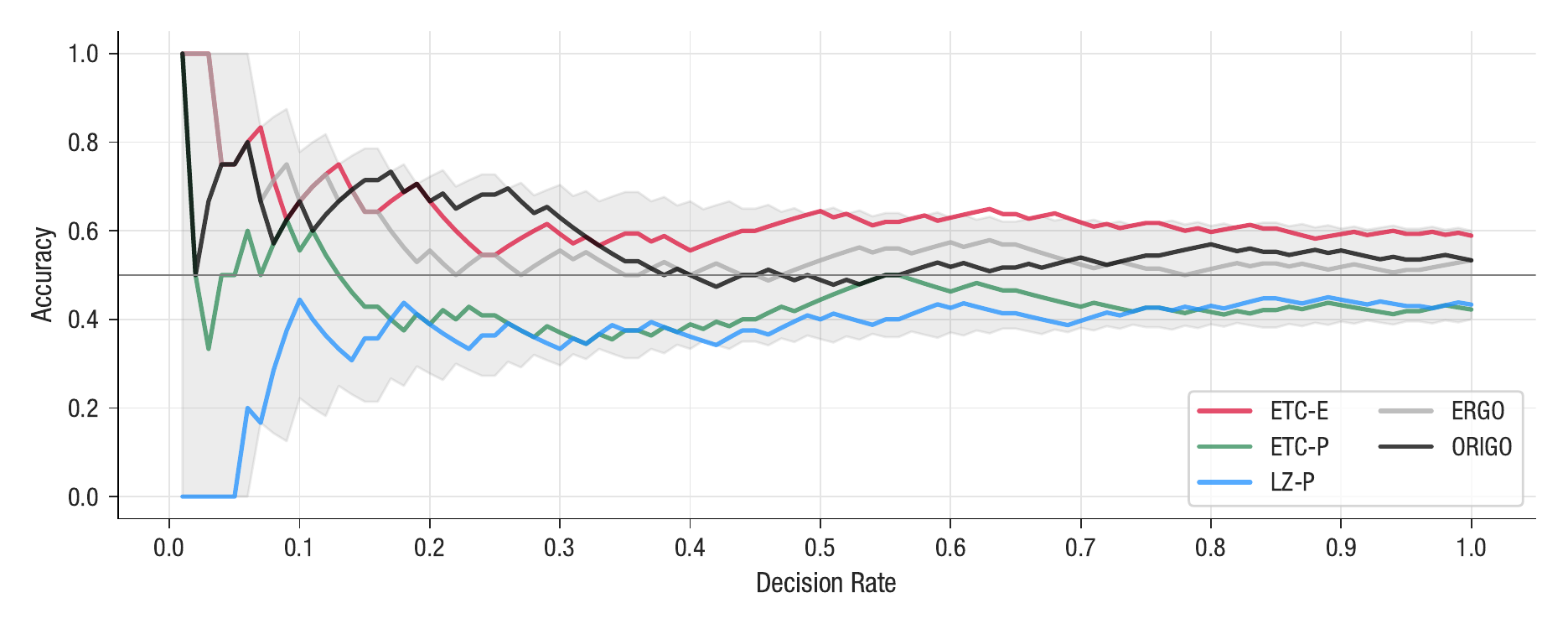}
	\caption{Accuracy of the five models across decision rates for the Tübingen cause-effect pairs. Gray band indicates 95\% CI of the binomial distribution.}
	\label{FIG:4}
\end{figure*}

\begin{figure}
	\centering
		\includegraphics[scale=0.95, center]{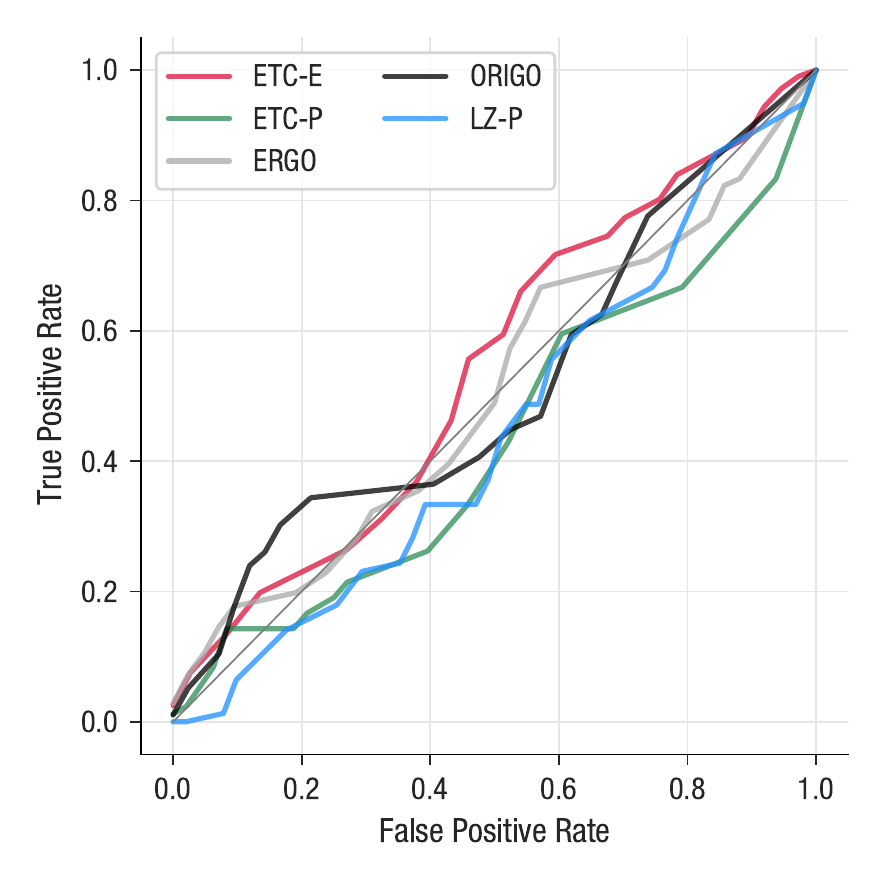}
	\caption{Receiver Operating Characteristic (ROC) curves for the five models for the Tübingen cause-effect pairs.}
	\label{FIG:5}
\end{figure}

\begin{table}
\caption{Area under the Receiver Operating Characteristic curve (AUROC) and Area under the Precision-Recall Curve (AUPRC) for the five models for the Tübingen cause-effect pairs}\label{tbl1}
\begin{tabular}{lllll}
\toprule
Model & AUROC & AUPRC \\
\midrule
ETC-E & 0.545 & 0.651 \\
ETC-P & 0.514 & 0.459 \\
LZ-P & 0.451 & 0.399 \\
ERGO & 0.504 & 0.590 \\
ORIGO & 0.524 & 0.591 \\
\bottomrule
\end{tabular}
\end{table}

\begin{figure*}
\centering
	\includegraphics[scale=0.95, center]{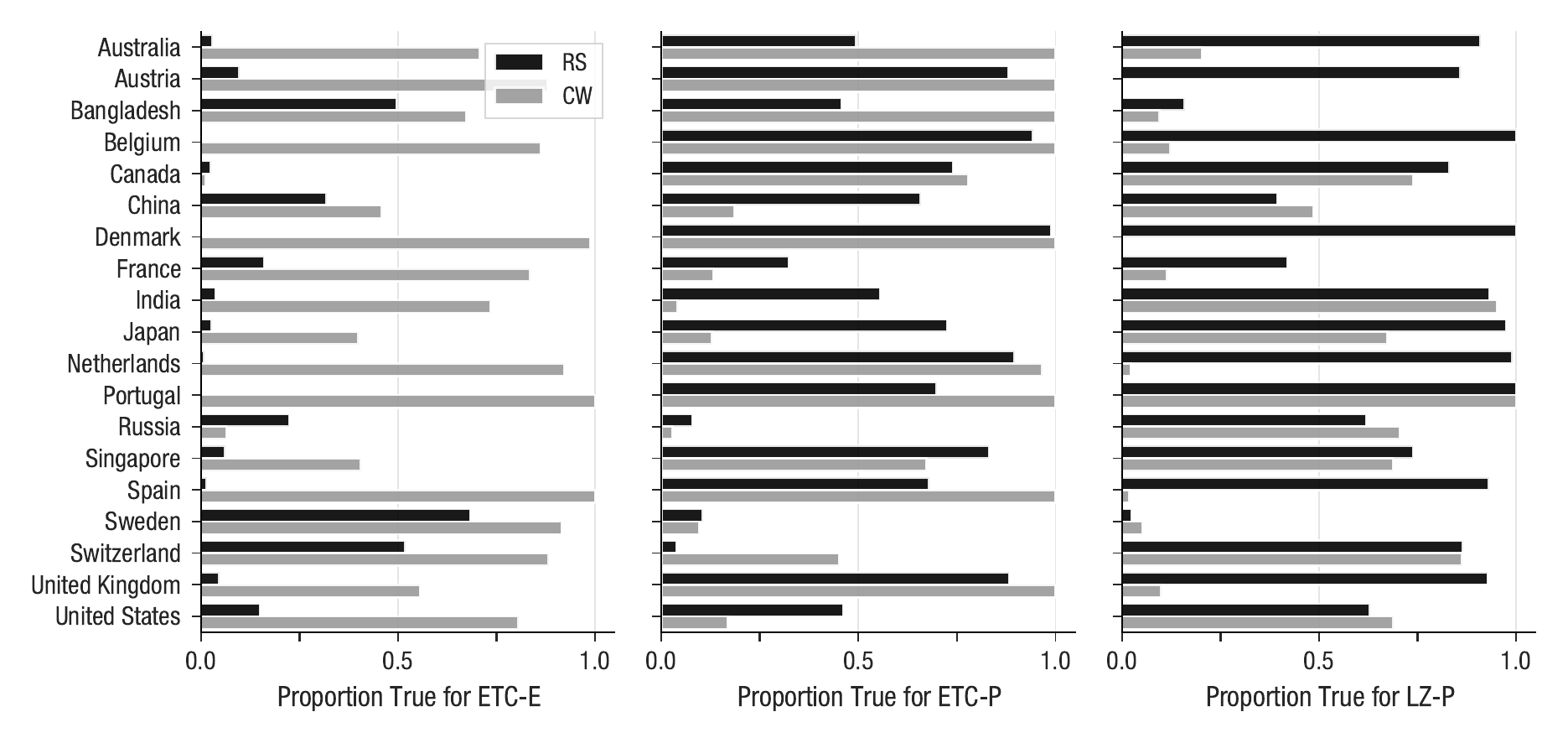}
	\caption{Proportion of sequences for which causal influence was discovered from either the RefSeq (RS) or the first sequence reported for that country (CW) under each of the three models - ETC-E, ETC-P, LZ-P.}
\label{FIG:6}
\end{figure*}

\begin{figure*}
\centering
	\includegraphics[scale=0.95, center]{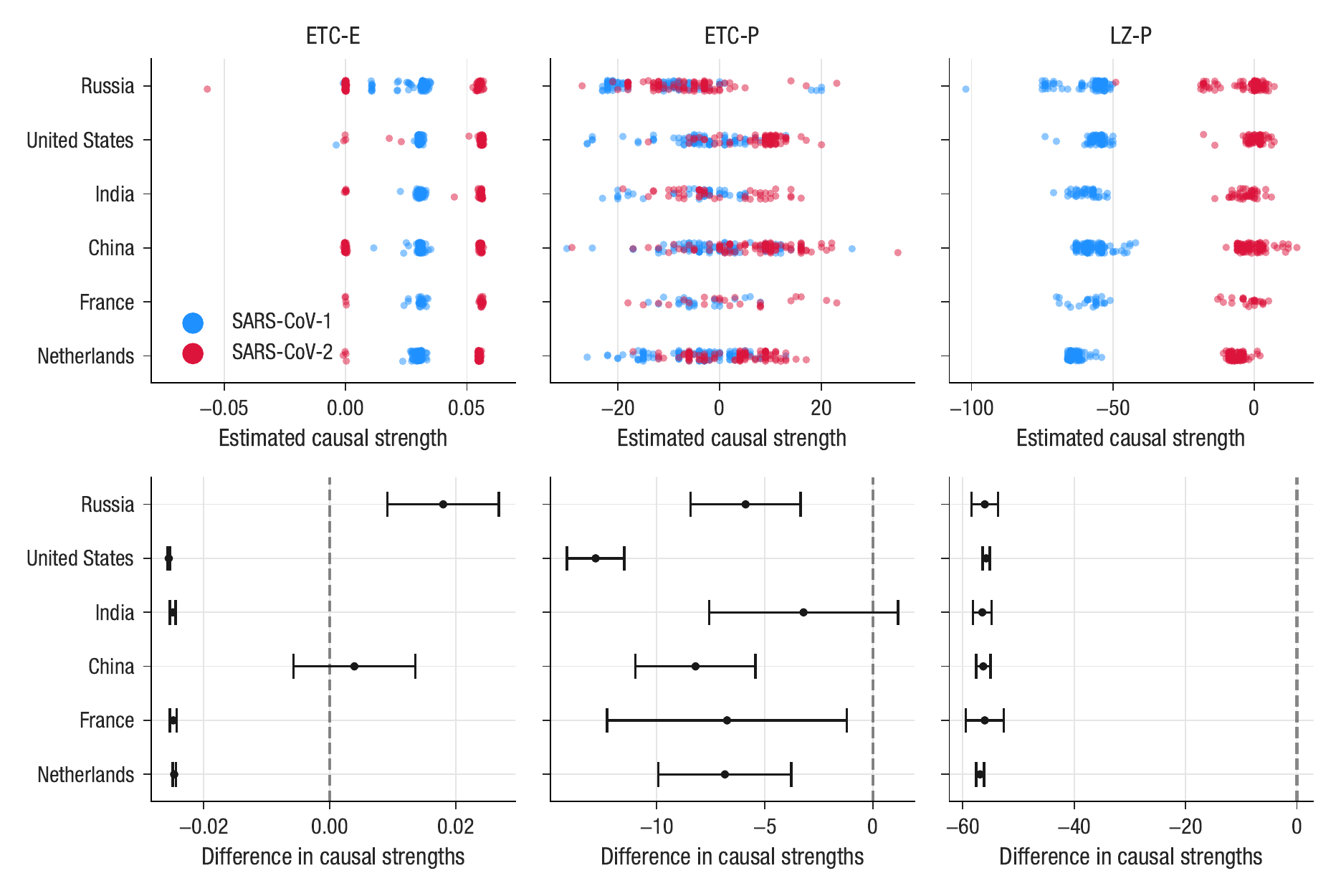}
	\caption{\textit{Top row:} Estimated strengths of causal influence from SARS-CoV-1 and SARS-CoV-2 for each sequence obtained from the three models - ETC-E, ETC-P, LZ-P. \textit{Bottom row:} Estimated differences in strengths of causal influence for each model expressed as the 20\% trimmed mean of causal strengths from SARS-CoV-1 $-$ 20\% trimmed mean of causal strengths from SARS-CoV-2. Error bars indicate bootstrapped 95\% confidence intervals around the trimmed means.}
\label{FIG:7}
\end{figure*}

\section{Results}
All experiments were executed in parallel across multiple cores on a workstation with a 1.8GHz 16-core Intel Xeon Silver 4108 CPU and 32GB of memory, running 64-bit Ubuntu 16.04LTS on kernel version 4.4.

\subsection{Synthetic unidirectional coupling}
With varying strengths of coupling for coupled AR processes, we observe in Fig. \ref{FIG:1} that the accuracy of LZ-P is nearly 90\% even at a low coupling strength of 0.2, improves dramatically and reaches 100\% accuracy for coupling strength beyond 0.35. The accuracies of ETC-P, ETC-E and ERGO improve slightly with coupling while that of ORIGO remains constant. Estimates of accuracy for each value of coupling strength are derived from 1,000 trials. The average accuracy obtained for coupling strengths greater 0.3 is highest for LZ-P at 98\%, followed by ERGO, ETC-E and ETC-P at 72\%, 61\% and 58\% respectively.

We examine the performance of all models as accuracy against the fraction of decisions each model is forced to make, or decision rates. As shown in Fig. \ref{FIG:2}, accuracies for LZ-P improve with decision rates for low coupling and are consistently high for stronger coupling. Similar to the observations from Fig. \ref{FIG:1}, the performance of LZ-P converges rapidly to 100\% accuracy near the coupling strength of 0.3. With stronger coupling overall accuracies improve with decision rates for ETC-E, ETC-P and ERGO, while they worsen slightly for ORIGO.

We further investigate the performance of all models using estimates of AUROC and AUPRC, summarized across varying coupling strengths in Fig. \ref{FIG:3}. We observe a rapidly increasing trend in AUROC estimates for LZ-P with coupling strengths, indicating that larger values of causal strengths obtained from LZ-P tend to favor the true causal direction. This performance saturates above coupling strengths of 0.5. While AUROC for ETC-P, ETC-E and ERGO are approximately constant and slightly better than chance for coupling strengths greater than 0.3. AUPRC estimates reveal that ETC-E, ETC-P and ERGO are able to correctly identify the true direction with similar performance, with LZ-P significantly outperforming the other models. ORIGO had AUROC and AUPRC estimates of less than 0.5 across all coupling strengths.

\subsection{Tübingen Cause-Effect Pairs}
We investigate the accuracy of all five models against the fraction of decisions each model is forced to make. In Fig. \ref{FIG:4}, we show the accuracy versus the decision rate for the benchmark Tübingen cause-effect pairs. If we look at all the pairs, we observe that ETC-E infers the correct direction roughly 60\%, followed by 54\% for ORIGO and ERGO. Both LZ-P and ETC-P infer the correct direction for only 42\% of all pairs. Considering only those pairs for which the strength of causal evidence was relatively large, we observe that over the top 8\% most decisive pairs, ETC-E is 78\% accurate and 70\% accurate for the top 20\% pairs. 

ROC curves for all models shown in Fig. \ref{FIG:5} suggest that causal strength is not very reliable at prediction of causal direction and performance is only slightly better than random. Large estimates of causal strengths do not necessarily correspond to the true direction. Estimates of AUROC, summarized in Table \ref{tbl1}, reflect the same. We do observe that of all models, AUROC as well as AUPRC are highest for ETC-E, followed by ORIGO, however, the performance of these two models does not compare favorably with the state-of-the-art causal inference frameworks for the Tübingen cause-effect pairs which frequently exceed 80\% overall accuracy on this dataset. \cite{mooij_distinguishing_2015, budhathoki_origo_2018, marx_telling_2019, vreeken_causal_2015}

\subsection{SARS-CoV-2 genome sequences}
We tested novel applications of the proposed models for causal inference of the direction and strength of information flow between genomic sequences of the SARS-CoV-2 virus. In experiment 1, we evaluated three hypotheses regarding causal discovery in genomic sequences using the ETC-E, ETC-P and LZ-P models.

Firstly, we assessed whether the SARS-CoV-2 consensus sequence (RS) 'causes' all other sequences. 9 countries for ETC-E and 18 countries for both ETC-P and LZ-P out of 19 countries had at least 5\% sequences which admitted this hypothesized direction. Of the three models, higher proportions across countries were generally observed under the ETC-P and LZ-P models.

Secondly, we assessed whether the first SARS-CoV-2 genome sequence isolated in each country (CW) 'causes' all other sequences isolated in that country. 18 countries for ETC-E, 17 for ETC-P and 16 for LZ-P out of 19 countries had at least 5\% sequences which admitted this hypothesized direction. Of the three models, higher proportions across countries were again generally observed under the ETC-P and LZ-P models.

Lastly, we assessed whether the proportions of sequences per country 'caused' by CW were greater than RS. This was the case with 17 countries for ETC-E, 11 for ETC-P and 5 for LZ-P out of 19 countries. Proportions per country obtained under each model are presented in Fig. \ref{FIG:6} and estimates provided in supplement S4.

In experiment 2, we examined whether estimated causal strengths from the proposed models could be used to identify a causal sequence from candidate sequences. We hypothesized that strength of causal influence on SARS-CoV-2 genome sequences would be greater from the SARS-CoV-2 consensus sequence than the SARS-CoV-1 consensus sequence.

In Fig. \ref{FIG:7}, individual estimates of strength from each of the three models for six countries are presented in the top row. Separation of the distribution of causal strengths is greatest for LZ-P across all countries, followed by ETC-E and then ETC-P. We evaluated this separation statistically for each country and model. In the bottom row of Fig. \ref{FIG:7}, we present the estimated differences in causal strengths as (strength from SARS-CoV-1) $-$ (strength from SARS-CoV-2) with bootstrapped 95\% confidence intervals. We observe that for the LZ-P case, causal strength was significantly lower for SARS-CoV-1 across all countries. For ETC-P, this difference was significantly lower for all countries except India, while for ETC-E, this difference was significantly lower for all countries except Russia and China. Note that for ETC-E, a distinct cluster of nearly 0 causal strengths is present for SARS-CoV-2, which, although not done here, can be appropriately thresholded. Complete statistical results are reported in supplement S3.

\section{Discussion}
Our experiments demonstrate that the three proposed models, LZ-P, ETC-P and ETC-E are able to infer causal direction adequately in practice. On simulated coupled autoregressive processes, LZ-P achieves a very high accuracy across coupling strengths, with 100\% accuracy for coupling strengths greater than 0.35. Both ETC-P and ETC-E models perform well for moderate to high coupling strengths on synthetic tests, while on benchmark data they perform at least as well as the compared models. We observed this performance despite the fact that both these experiments involved information loss due to discretization of continuous real-valued data. We attribute this performance to effective inference of unique context-free grammars (CFGs) by LZ and ETC from input symbolic sequences. The inferred grammars likely robustly capture information content of sequences, permitting an effective assessment of information shared and exchanged between sequences by the three models.

The systematically high performance of LZ-P in synthetic experiments suggests an alternate interpretation of the proposed model itself. We argue that due to the algorithmic differences between LZ and ETC, the estimated conditional CCMs (CCM$_L(y|G_x)$ and CCM$_L(x|G_y)$) may be capturing different information structures. In the ETC-P case, the term CCM$_L(y|G_x)$ decomposes into ETC$(y|G_x)$ and ETC$(y_{\text{residual}})$, where the former term corresponds to the cross-compression or conditioning step. This step halts when the inferred grammar $G_x$ can not compress $y$ further. However in the case of LZ-P, CCM$_L(y|G_x)$ in the presented work is estimated as LZ$(xy)$, where $xy$ is the sequence obtained by concatenation of $x$ and $y$. Here, CCM$_L(y|G_x)$ implicitly includes the CCM of $x$, leading to an interpretation in terms of conditioning analogous to conditional entropy. LZ-P likely captures the information accumulated jointly in $x$ and $y$, beyond information contained in $y$ itself (LZ$(y)$). The inference of direction now corresponds to lesser accumulation of information in the joint case compared to the information present in either $x$ or $y$. Further experiments are needed to identify conditions under which the LZ-P model can behave robustly and investigate the extent to which the suggested notions of `conditional' and `joint' LZ complexities can account for causal inference.

The cumulative entropy-based ERGO performs better than models other than LZ-P for synthetic data on the grounds of overall accuracy. Accuracy over top 20\% most decisive coupled pairs as well as AUPRC estimates highlight lack of differences in performance between ERGO, ETC-E and ETC-P. AUROC estimates on the other hand reveal an inadequate performance of ERGO, particularly for moderate to strong degrees of coupling. Differences can potentially arise due to the robustness of CCMs over entropy as complexity measures especially for short noisy time series, as well as due to the impact of conditioning. Under the two-part definition of Kolmogorov complexity used for ERGO, the cost of conditioning one sequence using the compressed representation of another sequence is always non-negative. While in the case of ETC-P, conditioning a sequence on the grammar of another sequence may result in either greater or poorer compression (CCM$_L(y|G_x)$ can be greater than, less than or equal to CCM$_L(y)$), resulting in positive or negative or zero penalties respectively, since $P_{x\rightarrow y}$ = CCM$_L(y|G_x)$ - CCM$_L(y)$. This likely allows ETC-P to capture different patterns of information exchange between sequences.

Our unique application of the proposed models for causal inference in SARS-CoV-2 genome sequences offers insights into directional information exchange between sequences. While each of the three models employed provide different causal perspectives, they all relied on inferred generative grammars underlying nucleotide sequences which uniquely characterize them. In the absence of constitutional temporal information in genome sequences, the proposed models were able to reliably infer a direction of causal information flow between pairs of sequences based on a plausible hypothesis. Similarly, causal strengths estimated from the models could robustly delineate differences in causal interactions from two different candidate sequences. To the best of our knowledge, such an application of grammar-based compression for examining causal interactions in nucleotide sequences has not been presented previously.

The results for causal inference are encouraging and showcase an application to genome sequences with very different lengths which does not require sequence alignment and examines sequences globally by decomposing their grammars. We believe these properties to be highly desirable for various applications in bioinformatics and systems biology such as estimating network graphs of sequence or gene interactions through pairwise causal inference, phylogenetics through the use of transformed strengths as distances, motif discovery based on an approach similar to the one presented in experiment 2 for quantifying dependence and interactions between sequences and regions on sequences, among others \cite{rao_motif_2007, aktulga_identifying_2007, conery_aligning_2008, nalbantoglu_data_2009}. These applications may be beneficial to epidemiological studies based on nucleotide sequences that rely on temporal information such as contact-tracing, founder effect, epidemic monitoring, evolution of virulence and pathogenicity \cite{salvamani_understanding_2020, ruan_founder_2020}.

While our experiments spanned synthetic, benchmark and real world data, a more extensive evaluation of performance of the proposed models is needed. In addition to linear AR coupling, linearly as well as non-linearly coupled dynamical systems, such as skew-tent maps, may characterize model performance with greater detail. Robustness to various kinds of noise and scaling, as well as applications to various publicly available benchmark data sets may help elucidate differential robustness, if any, of each of the models. The ETC-E model could particularly benefit from investigations into a more suitable normalization factor, $\lambda$, which may yield a better and more consistent performance for causal inference. The overall performance of all models tested is ultimately impacted by the choice of a discretization strategy, and while this remains an open problem, the usage of different methods such as K-Nearest Neighbours (KNNs) \cite{altman_introduction_1992}, Symbolic Aggregate approXimation (SAX) \cite{lin_experiencing_2007}, etc, may be considered definite improvements. 

\section{Conclusions}
We presented a new grammar-based information theoretic framework for univariate causal discovery using Compression-Complexity Measures (CCMs) and proposed two distinct approaches based on this framework. The penalty and efficacy approaches estimate the cost and effectiveness, respectively, of compressing a sequence using a grammar other than its own. We use these estimates in both directions for a pair of sequences to identify a causal direction based on \textit{less penalty} and \textit{greater efficacy} respectively.

For application in practice, we implemented these models using the lossless compressors ETC and LZ which serve as CCMs. The three models - ETC-P, ETC-E and LZ-P allow for reliable causal inference without demanding temporal structures of the data or making any assumptions about the data or its distributions. Empirical evaluation showed that the models are very accurate and perform competitively when compared to the state-of-the-art methods for synthetic benchmarks. Further, the fact that these models are essentially parameter-free makes their application in practice very easy compared to methods like Transfer Entropy \cite{schreiber_measuring_2000}, or CCC \cite{kathpalia_data-based_2019} which require careful parameter selection.

We presented a novel application of these models to causal analysis of SARS-CoV-2 genome sequences and found encouraging results which we believe can offer further insights into the disease transmission and sequence evolution. Our models are capable of discovering causal interactions directly from sequences, presenting opportunities for targeting key issues in bioinformatics, genomics and systems biology. As future work, we intend to refine the models presented, evaluate their performance on different causal interactions, and extend them for inference of causal networks.

\section*{Acknowledgement}
The authors gratefully acknowledge the financial support of `Cognitive Science Research Initiative' (CSRI-DST) Grant No. DST/CSRI/2017/54(G), `Science and Technology of Yoga and Meditation' (SATYAM-DST) Grant No. DST/SATYAM/2017/45(G) and the Tata Trusts towards this research.

\bibliographystyle{unsrt}  
\bibliography{references}

\end{document}